\documentclass[conference]{IEEEtran}
\IEEEoverridecommandlockouts

\usepackage{longtable}

\usepackage{booktabs}
\usepackage{float}
\usepackage{algorithm2e}

\usepackage{cite}
\usepackage{amsmath,amssymb,amsfonts}
\usepackage{algorithmic}
\usepackage{graphicx}
\usepackage{textcomp}
\usepackage{xcolor}
\usepackage{float}
\usepackage{subfigure}
\usepackage{multirow}
\usepackage{array}
\def\BibTeX{{\rm B\kern-.05em{\sc i\kern-.025em b}\kern-.08em
    T\kern-.1667em\lower.7ex\hbox{E}\kern-.125emX}}
\begin{document}

\title{Inverted Non-maximum Suppression for more Accurate and Neater Face Detection
}
\author{\IEEEauthorblockN{Lian Liu\IEEEauthorrefmark{1}\IEEEauthorrefmark{2}, Liguo Zhou\IEEEauthorrefmark{2}}
\IEEEauthorblockA{\IEEEauthorrefmark{1}College of Electronic and Information Engineering, Tongji University, Shanghai, China \\
\textit{1933424@tongji.edu.cn}
}
\IEEEauthorblockA{\IEEEauthorrefmark{2}Chair of Robotics, Artificial Intelligence and Real-time Systems, Technical University of Munich, Garching, Germany \\
\textit{liguo.zhou@tum.de}
}
}






%
\maketitle
\begin{abstract}
CNN-based face detection methods have achieved significant progress in recent years. In addition to the strong representation ability of CNN, post-processing methods are also very important for the performance of face detection. In general, the face detection method predicts several candidate bounding-boxes for one face. NMS is used to filter out inaccurate candidate boxes to get the most accurate box. The principle of NMS is to select the box with a higher score as the basic box and then delete the box which has a large overlapping area with the basic box but has a lower score. However, the current NMS method and its improved versions do not perform well when face image quality is poor or faces are in a cluster. In these situations, even after NMS filtering, there is often a face corresponding to multiple predicted boxes. To reduce this kind of negative result, in this paper, we propose a new NMS method that operates in the reverse order of other NMS methods. Our method performs well on low-quality and tiny face samples. Experiments demonstrate that our method is effective as a post-processor for different face detection methods. The source code has been released on https://github.com/.
\end{abstract}
\section{Introduction}

Face detection is an important task of computer vision and has been widely studied in the past decades. Nowadays, many emerging applications, such as security surveillance and identity authentication, hinge on face detection. As a special kind of object detection, the progress in face detection benefits from the developments in general object detection. The idea of object detection is to build a model with some fixed set of classes we are interested in. When an object belonging to a class appears in the input image, the bounding box is drawn around that object along with predicting its class label. The traditional stage was around 2000. Most of the methods proposed during this period were based on sliding windows and artificial feature extraction, which had the defects of high computational complexity and poor robustness in complex scenarios. Representative achievements include Viola-Jones detector~\cite{robust} and HOG pedestrian detector~\cite{histograms}. The second stage is from 2014 to the present, starting with the R-CNN~\cite{rich} proposed in 2014. These algorithms use Convolutional Neural Network (CNN)~\cite{lenet} to automatically extract hidden features in input images and classify and predict samples with higher accuracy. After R-CNN, there are many object detection methods based on CNN such as Fast R-CNN~\cite{fast}, Faster R-CNN~\cite{faster}, SSD~\cite{ssd}, and YOLO series~\cite{you}~\cite{yolov3}~\cite{yolov5}. Compared with the traditional object detection methods, the object detection methods based on CNN have the characteristics of high speed, strong accuracy, and high robustness.

\begin{figure}[H]
    \centering
    \includegraphics[width=8.6cm]{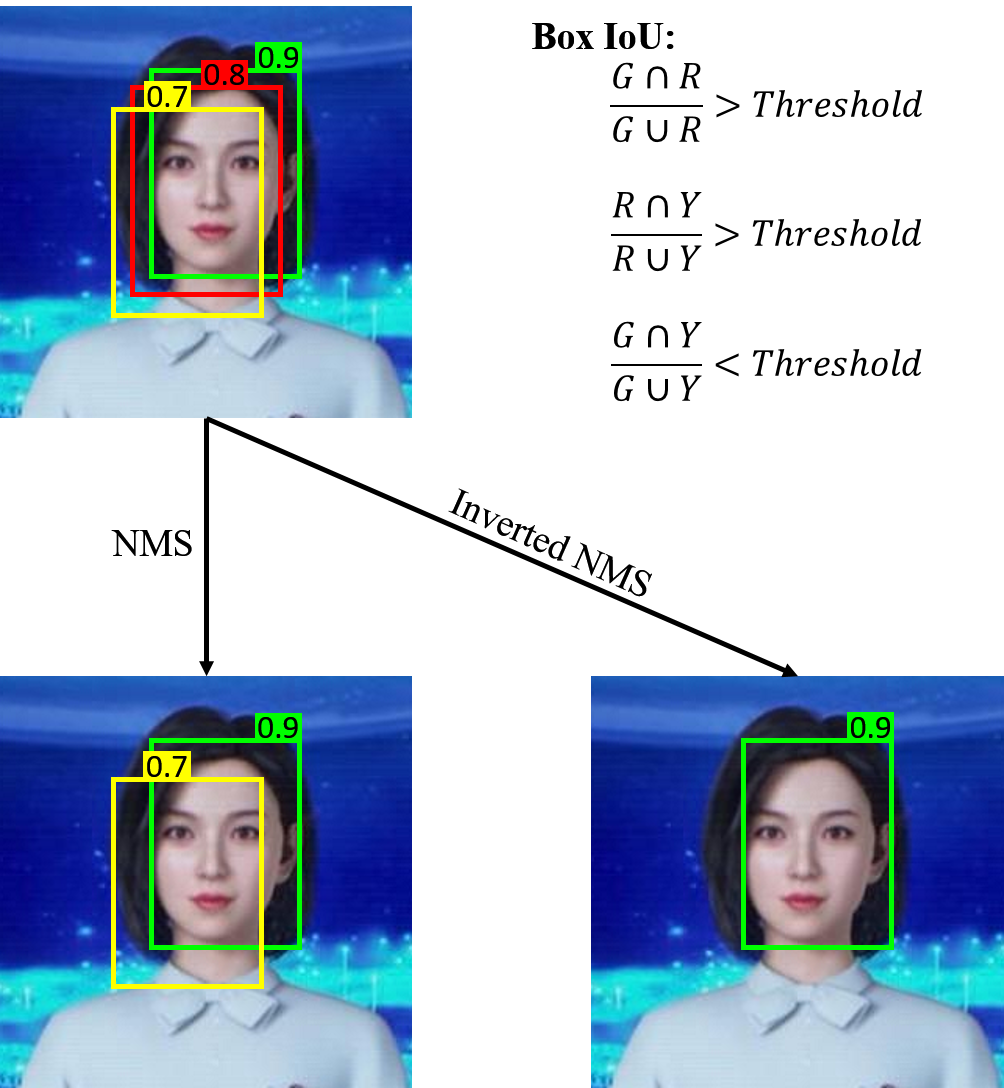}
    \caption{Three bounding-boxes, Green(G), Red(R) and Yellow(Y), are produced by a face detection method. The scores for G, R and Y are 0.9, 0.8 and 0.7, respectively. Post-processing by our Inverted NMS can get a better and neater result.}
    \label{fig 1}
\end{figure} 

In the test phase, CNN-based detection models output a large number of candidate bounding-boxes which contain a lot of redundancy. The CNN model also gives each box a score indicating the confidence that it surrounds an object correctly. Non-maximum suppression (NMS) is a commonly used post-processing method for discarding redundant predicted bounding-boxes. NMS is an iterative method to preserve local maximum and remove local non-maximum. In NMS, the candidate boxes are arranged in a list by sorting their scores in descending order. Then the box with the highest score is picked for calculating the Intersection over Union (IoU) values between it and all the other boxes. If an IoU value is larger than the pre-set threshold, the corresponding box with 
lower scores is deleted from the list. The picked box is also removed from the list and saved as a final box. The above process is repeated for the remaining list until the list is empty. As shown in Fig.~\ref{fig 1}, the Green box will definitely be preserved because it has the highest score. According to the above process of NMS, the Yellow box will also be preserved because the IoU between the Green box and Yellow box is less than the threshold and the Red box has been deleted before calculating the IoU between it and the Yellow box.

The disadvantage of NMS is obvious as shown in Fig~\ref{fig 1} and this situation is common in practical applications. Therefore, in this paper, we propose Inverted NMS to eliminate such shortcomings. Instead of arranging the candidate boxes by sorting their scores in descending order, we arrange a candidate box list in ascending order. Then we pick the box with the lowest score and calculate the IoU values between it and all the other boxes. If one of the IoU values is larger than the threshold, we delete the picked box and then repeat the progress above. Finally, the rest boxes in the list are the results of our Inverted NMS. As shown in Fig~\ref{fig 1}, according to our Inverted NMS, the Yellow box is deleted first because the IoU value between it and the Red box is larger than the threshold. Then the Red box is deleted because the IoU value between it and the Green box is larger than the threshold. It is obvious that our method can achieve neater results and the experiment section demonstrates that our method can improve the performance of detection on hard and tiny face samples.

\section{Related Work}

NMS has very important applications in the field of computer vision. In edge detection~\cite{computational}, after calculating the gradient value and gradient direction, the amplitude value is suppressed along the gradient direction by non-maximum value, and the irrelevant points that do not constitute an edge are removed, so that the possibility of it being an edge is excluded. In face detection, Cascade CNN~\cite{convolutional} uses Non-Maximum Suppression (NMS) to merge highly overlapping detection windows, and the remaining candidate detection windows will be normalized to 24$\times$24 as the input of 24-net, which will further eliminate the remaining nearly 90\% detection windows. In object detection, Faster R-CNN~\cite{faster} uses NMS in the proposal stage, the purpose is to remove the proposals that predict the same area with more serious overlap, keeping only the proposals with higher confidence. In the test phase of R-CNN, NMS is used for removing the low scored boxes that are overlapped with high score boxes.

NMS has a potential disadvantage of manually set threshold. Several alternatives have been considered. Some improved NMS methods are based on learning method. For instance, ConvNMS~\cite{convnet} is used to solve the difficult problem of NMS setting in the threshold. If the IoU threshold is set too high, the suppression may not be sufficient, some unnecessary predicted bounding-boxes may still be kept. If the IoU threshold is set too low, multiple true positives may be merged together. ConvNMS designs a convolutional network to combine the NMS results with different overlap thresholds and obtains the best output through the learning method. However, retraining and parameter tuning should be required in order to be effective in different scenarios. For the special application scenario of pedestrian detection in crowd, adaptive-NMS~\cite{adaptive} applies a dynamic suppression strategy, the suppression threshold in the instance is dynamically changed according to the target density, so that in densely crowded places, the NMS threshold is larger to obtain higher Recall, and where the crowd is sparse, NMS chooses a small threshold to eliminate more redundant boxes.

Some improved approaches for NMS include non-training procedures to progressively remove redundant bounding-boxes. Soft-NMS~\cite{soft} is a generalization of Traditional NMS, which is mainly aimed at alleviating the violent elimination of Traditional NMS. Soft-NMS introduces a re-scoring function, If the IoU is larger, the impact on score Si will be greater and Si will be smaller. In this way, the value of Si of each Box is updated, and the remaining Si, which is greater than a confidence threshold value, is retained to filter out candidate boxes. The Soft-NMS algorithm has improved on the standard datasets PASCAL VOC2007 (1.7\% higher than R-FCN and Faster-RCNN) and MS-COCO (1.3\% higher than R-FCN, 1.1\% higher than Faster-RCNN). This iterative procedure is friendly to two-stage methods, but it may fail in some single-stage methods. 

In Weighted NMS~\cite{inception}, the authors propose that the maximum score box selected by traditional NMS in each iteration may not be precisely positioned, and redundant boxes may also be well positioned. Weighted NMS is different from the direct culling mechanism, as its name implies, it is a weighted average of coordinates, and the objects of weighted average include instance in box set itself and adjacent boxes with IoU greater than NMS threshold. Weighted NMS usually achieves higher Precision and Recall, although the computational efficiency is lower than traditional NMS.

\section{Method}

\begin{algorithm}[h]
\caption{Inverted NMS} 
\label{Algorithm}
\KwIn{
$B = \{ b_1,b_2,...,b_n \}$, $S = \{ s_1,s_2,...,s_n \}$, $N_t$\\
\qquad $B$ is a set of predicted bounding boxes \\
\qquad $S$ is the corresponding predicted scores of $B$\\
\qquad $s_1 \le s_2 \le ... \le s_n$\\
\qquad $N_t$ is the NMS threshold
} 
\KwOut{$B^{\prime}$}
$D \gets \{\}$\\
\For{$i = 1; i\leq n-1; i++$}{
    \For{$j = i+1; j\leq n; j++$}{
        \If{$IoU(b_i, b_j) \ge N_t$}{
            $D \gets b_i$;\\
            Break;
        }
    }
}
$B^{\prime} \gets B-D$
\end{algorithm}

For one image, a CNN-based detection method usually outputs a large number of candidate bounding-boxes and each bounding-box has a score indicating the confidence that it contains a face correctly. In common, as shown in Fig.~\ref{fig 1}, a face may correspond to many bounding-boxes. Among them, some bounding-boxes are good while some bounding-boxes are bad. To remove the bad ones, we first arrange the candidate bounding-boxes by sorting their scores in ascending order. Then from top to bottom, we select boxes one by one and calculate the IoU values between the selected box and the boxes below it. If the IoU between the selected box and one of the boxes below is larger than a threshold, we delete the selected box. The detailed process is described in Algorithm~\ref{Algorithm}.

Our method relies heavily on the calculation of IoU. We describe the detailed calculation process below. Set the coordinates of two bounding boxes as $b_1(x_1,y_1,x_2,y_2)$ and $b_2(x_1^{'}, y_1^{'}, x_2^{'}, y_2^{'})$, where $(x_1,y_1)$ and $(x_1^{'}, y_1^{'})$ are the upper-left corners and $(x_2,y_2)$ and $(x_2^{'}, y_2^{'})$ are the lower-right corners. The area $a_1$ of $b_1$ and the area $a_2$ of $b_2$ can be obtained by 
\begin{equation}
\begin{split}
a_1&=(x_2-x_1) \times (y_2 - y_1),\\
a_2&=(x_2{'}-x_1{'}) \times (y_2{'} - y_1{'}).\\
\end{split}
\label{equation:1}
\end{equation}
The intersecting area of the two boxes can be obtained by
\begin{equation}
\begin{split}
a_{inter} &= max\{0, [min(x_2, x_2^{'})-max(x_1, x_1^{'})]\}\\
&\times max\{0, [min(y_2, y_2^{'})-max(y_1, y_1^{'})]\}
\end{split}
\label{equation:2}
\end{equation}
The IoU value is
\begin{equation}
IoU(b_1,b_2)=\frac{a_{inter}}{a_1+a_2-a_{inter}}.
\end{equation}

\section{Experiments}

\subsection{Setup}
We select five state-of-the-art object/face detection methods, YOLOv3~\cite{yolov3}, YOLOv5~\cite{yolov5}, DSFD~\cite{dsfd}, PyramidBox~\cite{pyramidbox} and EXTD~\cite{extd}, as our face detectors. All the detectors are trained on the WIDER FACE~\cite{wider} dataset by PyTorch~\cite{pytorch}. WIDER FACE contains a large number of faces with a high degree of variability in scale, pose, and occlusion. The validation set of WIDER FACE are split into three subsets, easy, medium and hard, which contains 7,211, 13,319 and 31,958 faces, respectively. We compare our Inverted NMS with the original NMS which is described in R-CNN~\cite{rich}, Weighted NMS~\cite{inception} and Soft NMS~\cite{soft} to demonstrate the effectiveness of our method.

In NMS, the threshold used to determine whether a box should be removed typically varies between 0.3 and 0.7 in order to obtain the best results. In our experiments, we try each threshold for each NMS method to obtain the best performance. As a result, the threshold for soft NMS should be 0.3 and the threshold for the other methods should be 0.6.

\begin{table}[ht]
\caption{Detection Results on WIDER FACE Val Set.}
\renewcommand\arraystretch{1.5}
    \centering
    \begin{tabular}{@{\hspace{0pt}}m{2cm}<{\centering}@{\hspace{0pt}}|@{\hspace{0pt}}m{2.6cm}<{\centering}@{\hspace{0pt}}|@{\hspace{0pt}}m{1.3cm}<{\centering}@{\hspace{0pt}}|@{\hspace{0pt}}m{1.3cm}<{\centering}@{\hspace{0pt}}|@{\hspace{0pt}}m{1.3cm}<{\centering}@{\hspace{0pt}}}
    \hline
    \hline
    \multirow{2}{*}{\textbf{Detector}} & \multirow{2}{*}{\textbf{NMS Method}} & \multicolumn{3}{c}{\textbf{Average Precision}}\\
    \cline{3-5}
    &  & \textbf{Easy} & \textbf{Medium} & \textbf{Hard}\\
    \hline
    \multirow{5}{*}{YOLOv5} & NMS & 0.962 & 0.961 & 0.907\\
     & Weighted NMS & 0.962 & 0.961 & 0.906\\
     & Soft NMS-L & 0.962 & 0.961 & 0.905\\
     & Soft NMS-G & 0.961 & 0.959 & 0.901\\
     & \textbf{Inverted NMS} & 0.962 & 0.961 & \textbf{0.924}\\
    \hline

    \multirow{5}{*}{YOLOv3}& NMS & 0.964 & 0.956 & 0.894\\
     & Weighted NMS & 0.964 & 0.956 & 0.893\\
     & Soft NMS-L & 0.964 & 0.955 & 0.892\\
     & Soft NMS-G & 0.963 & 0.954 & 0.888\\
     & \textbf{Inverted NMS} & 0.964 & 0.956 & \textbf{0.911}\\
         \hline

    \multirow{5}{*}{DSFD}& NMS & 0.949 & 0.935 & 0.847\\
     & Weighted NMS & 0.949 & 0.935 & 0.847\\
     & Soft NMS-L & 0.949 & 0.935 & 0.849 \\
     & Soft NMS-G & 0.950 & 0.936 & 0.844\\
     & \textbf{Inverted NMS} & 0.950 & 0.937 & \textbf{0.856}\\
         \hline

    \multirow{5}{*}{EXTD}& NMS & 0.918 & 0.905 & 0.828 \\
     & Weighted NMS & 0.917 & 0.904 & 0.825\\
     & Soft NMS-L & 0.920 & 0.905 & 0.784 \\
     & Soft NMS-G & 0.920 & 0.904 & 0.782 \\
     & \textbf{Inverted NMS} & 0.918  & 0.905 & \textbf{0.832}\\
         \hline

    \multirow{5}{*}{PyramidBox}& NMS & 0.946 & 0.934 & 0.853\\
     & Weighted NMS & 0.948 & 0.936 & 0.851\\
     & Soft NMS-L & 0.948 & 0.937 & 0.854\\
     & Soft NMS-G & 0.947 & 0.936 & 0.846\\
     & \textbf{Inverted NMS} & 0.948 & 0.936 & \textbf{0.859}\\
    \hline
    \hline
    \end{tabular}
    \label{table 1}
\end{table}

\begin{table}[ht]
\caption{Results of YOLOv5 on Faces of Different Sizes (WIDER FACE Val Set).}
\renewcommand\arraystretch{1.5}
    \centering
    \begin{tabular}{@{\hspace{0pt}}m{2.9cm}<{\centering}@{\hspace{0pt}}|@{\hspace{0pt}}m{1.4cm}<{\centering}@{\hspace{0pt}}|@{\hspace{0pt}}m{1.4cm}<{\centering}@{\hspace{0pt}}|@{\hspace{0pt}}m{1.4cm}<{\centering}@{\hspace{0pt}}|@{\hspace{0pt}}m{1.4cm}<{\centering}@{\hspace{0pt}}}
    \hline
    \hline
    &\multicolumn{4}{c}{\textbf{Average Precision}}\\
    \hline
    \textbf{Longer Side of GT} & $\le$16 &(16, 64] &(64, 256]&$>$256\\
    \hline
    \textbf{Number of GT} & 16844 & 17793 & 4482 & 586 \\
    \hline
    NMS & 0.610 & 0.926 & 0.961 & 0.672\\
    Weighted NMS & 0.609 & 0.925 & 0.961 & 0.670\\
    Soft NMS-L & 0.610 & 0.924 & 0.961 & 0.672\\
    Soft NMS-G & 0.607 & 0.922 & 0.961 & 0.674\\
    \textbf{Inverted NMS} & \textbf{0.686} & \textbf{0.927} & 0.961 & 0.672\\
    \hline
    \hline
    \end{tabular}
    \label{table 2}
\end{table}

\begin{figure}
    \centering
    \subfigure[Inverted NMS]{
    \includegraphics[width=4cm]{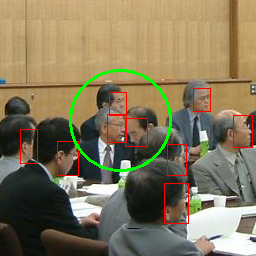}
    }
    \subfigure[Other NMS]{
    \includegraphics[width=4cm]{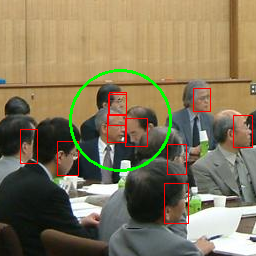}
    }
    \subfigure[Inverted NMS]{
    \includegraphics[width=4cm]{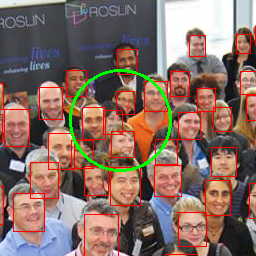}
    }
    \subfigure[Other NMS]{
    \includegraphics[width=4cm]{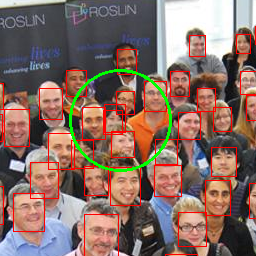}
    }
    \subfigure[Inverted NMS]{
    \includegraphics[width=4cm]{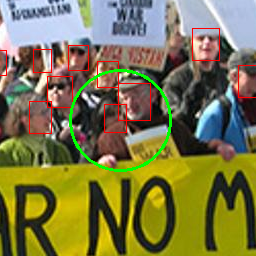}
    }
    \subfigure[Other NMS]{
    \includegraphics[width=4cm]{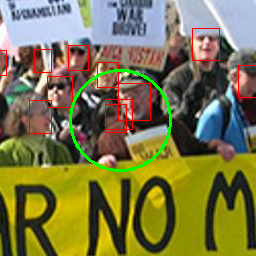}
    }
    \caption{Visualized results of different NMS methods. The reults of our method is neater.}
    \label{fig 2}
\end{figure}

\subsection{Results}
Table~\ref{table 1} shows the detection results of different detectors that combine different NMS methods. Our Inverted NMS can improve the performance of detectors on the hard subset of WIDER FACE validation set. Especially for YOLOv3 and YOLOv5, our method largely improves the performance of detecting hard face samples. Due to the high proportion of hard samples in the whole validation dataset, our Inverted NMS can improve the overall detection performance.

By counting the detection results on faces of different scales, we can find that our method is effective for detecting tiny faces. As shown in Table~\ref{table 2}, the detection performance of YOLOv5 with our Inverted NMS on tiny faces with side lengths less than 16 is significantly improved.

Fig.~\ref{fig 2} visualizes detection results of three face images. Compared with other NMS methods, our method has a good filtering performance on multiple boxes at some face clusters.

\subsection{Complexity Analysis}

In the original NMS, after completing a traversal, it is possible that multiple boxes will be deleted, which reduces the number of comparisons for the next traversal and the number of traversals. In our method, we delete at most one box per traversal, which means that our method will consume more time than the original method. However, the time consumption of our method is still milliseconds, which is negligible compared to the time consumption of the object detection network.

\section{Conclusion}
In this paper, we propose an Inverted NMS to eliminate the redundant predicted bounding-boxes surrounding hard face samples. Our method deletes the bad bounding-boxes by comparing the IoU start from the box with the lowest score while the other NMS methods start from the box with the highest score. The experiments demonstrate that our method is more effective than the others for detecting hard and tiny face samples.

\bibliographystyle{IEEEtran}
\bibliography{strings}

\end{document}